\documentclass[sigconf]{acmart}

\usepackage{booktabs} 

\usepackage{latexsym}
\usepackage{booktabs} 
\usepackage{algorithm}
\usepackage{algorithmic}
\usepackage{epsfig}
\usepackage{color}
\usepackage{flushend}
\usepackage{amsmath}
\usepackage{graphicx,subfigure}
\usepackage{multirow}
\usepackage{makecell}
\usepackage{subfigure}
\usepackage{CJKutf8}
\usepackage{textcomp}

\copyrightyear{2019}
\acmYear{2019} 
\setcopyright{iw3c2w3}
\acmConference[WWW '19]{Proceedings of the 2019 World Wide Web Conference}{May 13--17, 2019}{San Francisco, CA, USA}
\acmBooktitle{Proceedings of the 2019 World Wide Web Conference (WWW '19), May 13--17, 2019, San Francisco, CA, USA}
\acmPrice{}
\acmDOI{10.1145/3308558.3313437}
\acmISBN{978-1-4503-6674-8/19/05}

\usepackage{balance}

\begin{document}
\begin{CJK*}{UTF8}{gbsn}

\title{Neural Chinese Word Segmentation with Lexicon and Unlabeled Data via Posterior Regularization}


\author{Junxin Liu}
\affiliation{%
  \institution{Tsinghua University}
  \city{Beijing}
  \country{China}
}
\email{ljx16@mails.tsinghua.edu.cn}

\author{Fangzhao Wu}
\affiliation{%
  \institution{Microsoft Research Asia}
  \city{Beijing}
  \country{China}
}
\email{wufangzhao@gmail.com}

\author{Chuhan Wu}
\affiliation{%
  \institution{Tsinghua University}
  \city{Beijing}
  \country{China}
}
\email{wuch15@mails.tsinghua.edu.cn}

\author{Yongfeng Huang}
\affiliation{%
  \institution{Tsinghua University}
  \city{Beijing}
  \country{China}
}
\email{yfhuang@tsinghua.edu.cn}

\author{Xing Xie}
\affiliation{%
  \institution{Microsoft Research Asia}
  \city{Beijing}
  \country{China}
}
\email{Xing.Xie@microsoft.com}

\renewcommand{\shortauthors}{Junxin et al.}

\begin{abstract}
Chinese word segmentation (CWS) is very important for Chinese text processing.
Existing methods for CWS usually rely on a large number of labeled sentences to train word segmentation models, which are expensive and time-consuming to annotate.
Luckily, the unlabeled data is usually easy to collect and many high-quality Chinese lexicons are off-the-shelf, both of which can provide useful information for CWS.
In this paper, we propose a neural approach for Chinese word segmentation which can exploit both lexicon and unlabeled data.
Our approach is based on a variant of posterior regularization algorithm, and the unlabeled data and lexicon are incorporated into model training as indirect supervision by regularizing the prediction space of CWS models.
Extensive experiments on multiple benchmark datasets in both in-domain and cross-domain scenarios validate the effectiveness of our approach.
\end{abstract}

%
%



\maketitle

\section{Introduction}

Chinese word segmentation (CWS) aims to segment Chinese sentence into words~\cite{xue2003chinese,gao2005chinese,levow2006third}.
For example, ``习近平常与特朗普通电话'' is segmented into ``习近平/常/与/特朗普/通/电话''.
Different from English texts where whitespace is used to separate words, there is no natural word delimiter in Chinese.
Thus, CWS is very important for processing Chinese texts and is an essential step for many downstream tasks~\cite{luo2016empirical,chen2017adversarial,foo2004chinese}.

In recent years, neural network based methods have been widely used for CWS~\cite{zhang2016transition,yang2017neural,cai2017fast,pei2014max}.
Most of these methods model CWS as a sequence labeling problem~\cite{xue2003chinese,zhao2006effective}, and utilize neural networks to learn the hidden character features~\cite{zheng2013deep,chen2015long}.
For example, Chen et al.~\shortcite{chen2015long} used LSTM~\cite{hochreiter1997long} to learn character features by capturing the global information of sentence.
Peng and Dredze~\shortcite{peng2017multi} proposed to use LSTM for character feature learning and CRF~\cite{lafferty2001conditional} for character label decoding.
However, these methods usually rely on a large number of labeled sentences to train word segmentation models, which are expensive and time-consuming to annotate.
Besides, these methods usually have difficulty in segmenting sentences with OOV (out of vocabulary) words or words that are rare in training data~\cite{zhang2018neural}.
For example, if ``习近平'' and ``特朗普'' are OOV words in training data, then these methods will probably segment ``习近平常与特朗普通电话'' into ``习近/平常/与/特朗/普通/电话''.

Our work is motivated by following observations.
First, the unlabeled Chinese sentences are usually easy to collect on a large scale and can provide useful information for Chinese word segmentation.
For example, if the character sequences ``习近平'' and ``特朗普'' appear many times in unlabeled data with different contexts, then we can infer that they are probably Chinese words.
Second, many high-quality Chinese lexicons have been built and can cover a large number of Chinese words.
These lexicons can provide important information of whether a Chinese character sequence can be a valid Chinese word, which is useful for CWS.
For example, if ``习近平'' and ``特朗普'' are included in a Chinese lexicon, then we can better segment aforementioned sentences.
Thus, both unlabeled data and lexicons have the potential to improve the performance of CWS, especially on sentences with OOV and rare words.

In this paper, we propose a neural approach for Chinese word segmentation which can exploit the useful information in both Chinese lexicon and unlabeled data.
More specifically, in our approach we propose a unified framework based on posterior regularization~\cite{ganchev2010posterior} to incorporate Chinese lexicon and unlabeled data as indirect supervision to regularize the prediction space of the neural CWS models.
The neural CWS architecture used in our approach is CNN-CRF.
The neural CWS model is trained based on both indirect supervision inferred from lexicon and unlabeled data and the direct supervision inferred from labeled sentences.
Extensive experiments are conducted on multiple benchmark datasets in both in-domain and cross-domain scenarios.
The experimental results show that our approach can effectively improve the performance of Chinese word segmentation, especially when training data is insufficient.

\section{Related Work}

In recent years, neural network based methods have been widely used for Chinese word segmentation.
These methods usually regard CWS as a character-level sequence labeling task.
For example, Chen et al.~\shortcite{chen2015long} proposed to apply LSTM to Chinese word segmentation.
They used LSTM to learn hidden character features by capturing the global context information of sentences.
They also used a character window to capture local contexts for building character features.
Peng et al,~\shortcite{peng2017multi} used LSTM to learn the contextual character features and used CRF to jointly decode the labels of characters.
These neural methods for CWS usually rely on a large number of labeled sentences for model training.
When the labeled data is insufficient, the performance usually declines heavily~\cite{zhang2018neural,zhang2018addressing,xu2017transfer}.

Incorporating the useful information in unlabeled sentences and lexicons into CWS has attracted increasing attentions~\cite{li2009punctuation,sun2011enhancing,yang2017neural}.
For example, Li et al.~\shortcite{li2009punctuation} proposed to utilize unlabeled sentences for CWS by using punctuation marks as implicit annotations.
However, punctuation marks are sparse in Chinese texts, and the annotations of most characters cannot be obtained in this way.
Sun et al.~\shortcite{sun2011enhancing} proposed to extract statistics-based character features such as mutual information and accessor variety from unlabeled data, and use these features to improve CWS.
Designing these handcrafted features needs a large amount of domain knowledge.
Zhang et al.~\shortcite{zhang2018neural} proposed to incorporate lexicon into a neural CWS method based on LSTM-CRF architecture.
They designed several handcrafted feature templates to extract additional character features using lexicon, and used another LSTM to learn character representations from these lexicon based features.
Liu et al.~\shortcite{liu2018neural} proposed to utilize lexicon for neural CWS via multi-task learning.
They designed an auxiliary task of word classification to exploit lexicon information.
Then they jointly trained CWS and word classification models using a multi-task learning framework.
However, these methods cannot exploit the useful information in unlabeled data.

There are a few methods which can incorporate both lexicon and unlabeled data into Chinese word segmentation.
For example, Liu et al.~\shortcite{liu2014domain} proposed to build partially annotated data using unlabeled data and lexicons via many handcrafted rules. 
Then they trained CWS models based on both labeled and partially annotated data using CRF.
However, designing these rules requires a lot of domain knowledge.
Zhao et al.~\shortcite{zhao2018neural} used a similar way to build partially labeled data for CWS by combining unlabeled data and Chinese lexicons.
They trained neural CWS models on both labeled and partially labeled data using a variant of LSTM.
However, in these methods the partially annotated data is built simply by word matching without considering the contexts of sentences.
Thus, the partially annotated data may contain heavy noise and is not suitable to use directly as training data.
Different from these methods, in our approach the lexicon and unlabeled data are incorporated to provide indirect supervision via posterior regularization.

Posterior regularization algorithm~\cite{ganchev2010posterior} can incorporate additional knowledge into model training as constraints over the posterior distributions of models.
For instance, Zhang et al.~\shortcite{zhang2017prior} applied posterior regularization to integrate various kinds of knowledge such as bilingual dictionary, phrase table and coverage penalty into neural machine translation and gained huge performance improvements.
Our approach is motivated by posterior regularization, and we apply it to incorporate the useful information in Chinese lexicon and unlabeled data into neural Chinese word segmentation.

\section{Our Approach}

We first introduce the CNN-CRF neural architecture in our approach for Chinese word segmentation.
Then we introduce our approach to incorporate lexicon and unlabeled data into neural CWS.

\begin{figure}
  \centering
  \includegraphics[width=0.3\textwidth]{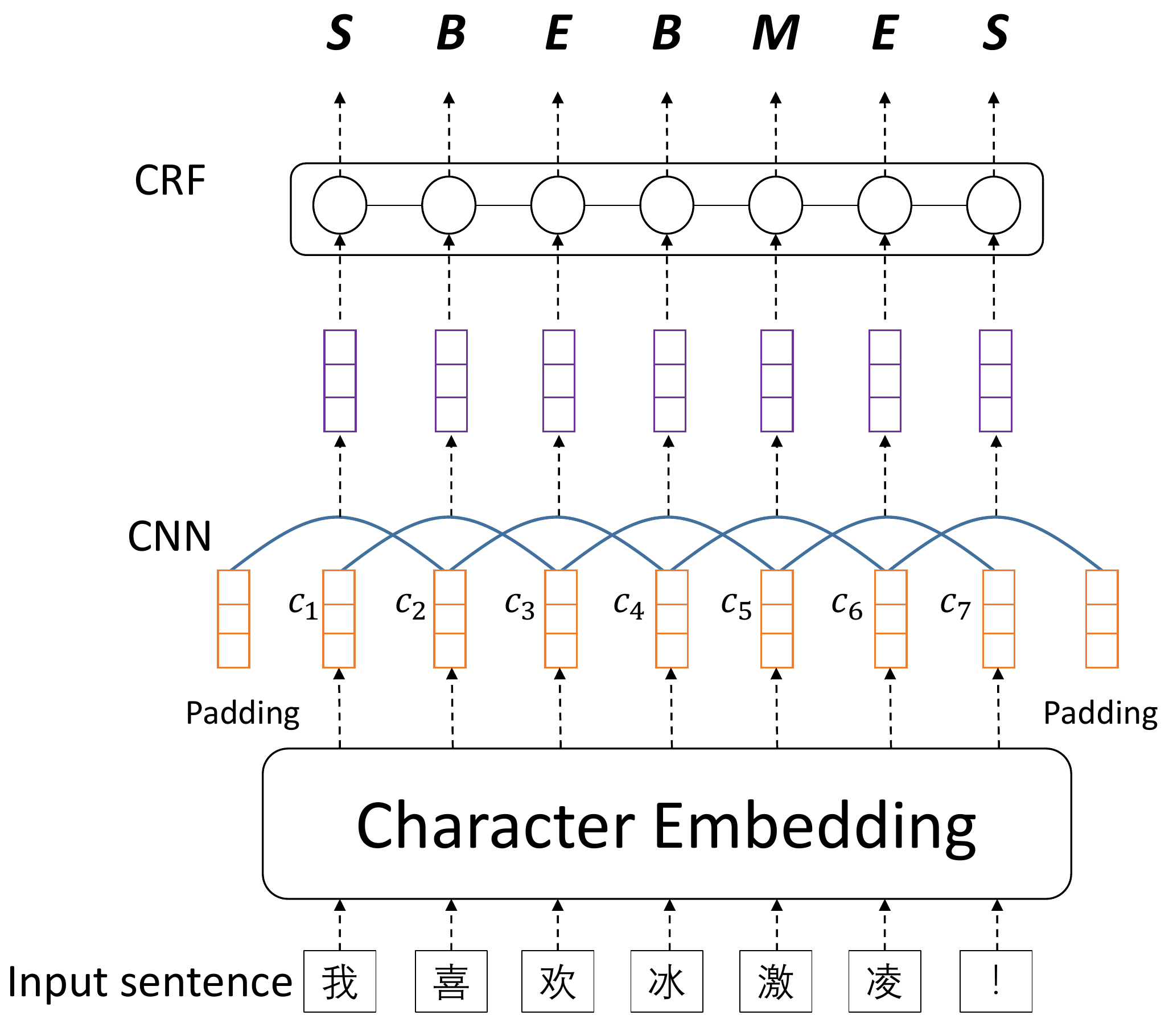}
  \caption{The CNN-CRF neural architecture for CWS.}
  \label{model}
\end{figure}

\subsection{CNN-CRF Architecture}

Following many previous works~\cite{low2005maximum,zhao2006improved}, we model Chinese word segmentation as a character-level sequence labeling problem.
For each character in a sentence, our model assigns a tag from a predefined tag set to it which indicates the position of this character in a word.
The tag set used in our model is $\{B,M,E,S\}$, where $B$, $M$ and $E$ mean the beginning, middle and end position in a word respectively, and $S$ means single character word.


We use CNN-CRF as the neural architecture for CWS which is illustrated in Fig.~\ref{model}.
The CNN-CRF architecture contains three layers.
The first one is character embedding.
Given a sentence $x=[c_1, c_2, ... , c_N]$, the character embedding layer will map each character to a low-dimensional dense vector.
Here $c_i$ is the $i$-th character in the sentence, and $N$ is the sentence length.
The output of this layer is $x=[\mathbf{e}_1, \mathbf{e}_2, ... , \mathbf{e}_N]$, where $\mathbf{e}_i$ is the embedding of $c_i$.

The second layer is a CNN network.
It is used to learn contextual representations of characters.
The hidden feature of the $i$-th character learned by a convolutional kernel is:
\begin{equation}
h_i=f(\mathbf{w}^T\times \mathbf{e}_{i-\lceil \frac{k-1}{2}\rceil:i+\lfloor \frac{k-1}{2}\rfloor}+b),
\end{equation}
where $\mathbf{w}$ and $b$ are the parameters of the convolutional kernel, $K$ is the kernel size, and $\mathbf{e}_{i-\lceil \frac{k-1}{2}\rceil:i+\lfloor \frac{k-1}{2}\rfloor}$ represents the concatenation of the embeddings from $i-\lceil \frac{k-1}{2}\rceil$-th character to $i+\lfloor \frac{k-1}{2}\rfloor$-th character.
We use multiple kernels with different kernel sizes and concatenate the outputs of these convolutional kernels as the feature representation for each character.
The output of CNN layer is $[\mathbf{h}_1,\mathbf{h}_2,...,\mathbf{h}_N]$, where $\mathbf{h}_i \in \mathcal{R}^{F}$, and $F$ is the number of kernels in the CNN network.

The third layer is CRF~\cite{lafferty2001conditional}.
Given a sentence $\mathbf{x}=[c_1, c_2, ... , c_N]$ and a tag sequence $\mathbf{y}=[y_1, y_2, ... , y_N]$, the score of sentence $\mathbf{x}$ having tag sequence $\mathbf{y}$ is formulated as follows:
\begin{equation}
s(\mathbf{x}, \mathbf{y})=\sum_{i=1}^{N}U_{i,y_i}+\sum_{i=1}^{N-1}A_{y_i,y_{i+1}},
\label{eq.segmentation_score}
\end{equation}
where $U_{i,y_i}$ represents the unary score of assigning tag $y_i$ to the $i$-th character, and $A_{y_i,y_{i+1}}$ represents the score of jumping from tag $y_i$ to tag $y_{i+1}$.
The unary score $\mathbf{U}_{i} \in \mathcal{R}^{T}$ is formulated as:
\begin{equation}
\mathbf{U}_{i}=\mathbf{W}_c\mathbf{h}_i+\mathbf{b}_c,
\end{equation}
where $\mathbf{W}_c\in \mathcal{R}^{T\times F}$ and $\mathbf{b}_c\in \mathcal{R}^{T}$ are trainable parameters, and $T$ is the size of the tag set.
Then the likelihood probability of sentence $\mathbf{x}$ having tag sequence $\mathbf{y}$ is defined as:
\begin{equation}
p(\mathbf{y}|\mathbf{x})=\frac{e^{\mathbf{s}(\mathbf{x},\mathbf{y})}}{\sum_{\mathbf{y'}\in \mathcal{Y}_{\mathbf{x}}}e^{\mathbf{s}(\mathbf{x},\mathbf{y'})}},
\end{equation}
where $\mathcal{Y}_{\mathbf{x}}$ represents the set of all possible tag sequences of sentence $\mathbf{x}$.
And the loss function is formulated as follows:
\begin{equation}\label{eq.loss_base}
\mathcal{L}(\boldsymbol{\theta})=-\sum_{i=1}^{N_l}\log (p(\mathbf{y}_i|\mathbf{x}_i;\boldsymbol{\theta})),
\end{equation}
where $\mathbf{x}_i$ is the $i$-th training sentence, $\mathbf{y}_i$ is its ground truth tag sequence, $N_l$ is the number of labeled sentences in training set, and $\boldsymbol{\theta}$ represents all parameters of the neural CWS model.

\subsection{Neural CWS with Lexicon and Unlabeled Data via Posterior Regularization}

In this section we introduce our approach to exploit lexicon and unlabeled data to train a neural CWS model.
Our approach is based on posterior regularization~\cite{ganchev2010posterior}, and we propose a unified framework to incorporate the useful information in lexicon and unlabeled data as indirect supervision into model training by regularizing the prediction space of neural CWS model.
In our approach the neural CWS model is trained in an iterative manner.
Following~\cite{zhang2017prior}, in iteration $t$, the loss function of the indirect supervision is:
\begin{equation}
\mathcal{L}^{PR}(\boldsymbol{\theta}) = \sum_{i=1}^{N_u}\text{KL}(Q(\mathbf{y}|\mathbf{x}_i;\boldsymbol{D},\boldsymbol{\theta}^t)||p(\mathbf{y}|\mathbf{x}_i;\boldsymbol{\theta})),
\label{eq.loss_PR}
\end{equation}
where $\boldsymbol{\theta}$ is the parameter set of CNN-CRF model, $\boldsymbol{\theta}^t$ is the model learned in iteration $t-1$, $\boldsymbol{D}$ represents Chinese lexicon, and $N_u$ is the number of unlabeled sentences.
$\text{KL}$ is the $\text{KL}$ divergence function.
$Q(\mathbf{y}|\mathbf{x};\boldsymbol{D},\boldsymbol{\theta}^t)$ is the probability distribution of tag sequence $\mathbf{y}$ for unlabeled sentence $\mathbf{x}$ given lexicon $\boldsymbol{D}$ and previous model $\boldsymbol{\theta}^t$:
\begin{equation}
Q(\mathbf{y}|\mathbf{x};\boldsymbol{D},\boldsymbol{\theta}^t)=\frac{\exp(\phi(\mathbf{y},\mathbf{x};\boldsymbol{D},\boldsymbol{\theta}^t))}{\sum_{\hat{\mathbf{y}}\in\mathcal{Y}(\mathbf{x})}\exp( \phi(\hat{\mathbf{y}},\mathbf{x};\boldsymbol{D},\boldsymbol{\theta}^t))},
\label{eq.Q1}
\end{equation}
where $\phi(\mathbf{y},\mathbf{x};\mathbf{D},\boldsymbol{\theta}^t)$ is a score function of tag sequence $\mathbf{y}$ for sentence $\mathbf{x}$, and $\mathcal{Y}(\mathbf{x})$ is the set of all possible tag sequences of $\mathbf{x}$. $\phi(\mathbf{y};\mathbf{x},\mathbf{D},\boldsymbol{\theta}^t)$ is designed to encode both lexicon information and the predictions of previous CWS model towards this sentence:
\begin{equation}
\phi(\mathbf{y};\mathbf{x},\boldsymbol{D},\boldsymbol{\theta}^t) = \frac{n(\mathbf{x}, \mathbf{y}; \boldsymbol{D})}{n(\mathbf{x}, \mathbf{y})}  + \alpha\cdot s(\mathbf{x}, \mathbf{y};\boldsymbol{\theta}^t),
\label{eq.phi}
\end{equation}
where $n(\mathbf{x}, \mathbf{y})$ is the number of words in the segmentation result, $n(\mathbf{x}, \mathbf{y}; \boldsymbol{D})$ is the number of words in segmentation result which are included in lexicon $\boldsymbol{D}$, $s(\mathbf{x}, \mathbf{y};\boldsymbol{\theta}^t)$ is the segmentation score predicted by the CWS model $\boldsymbol{\theta}^t$ trained in previous iteration according to Eq.~(\ref{eq.segmentation_score}), and $\alpha$ is a positive coefficient.
According to Eq.~(\ref{eq.phi}), if a tag sequence $\mathbf{y}$ for an unlabeled sentence $\mathbf{x}$ can lead to more lexicon-included words and has higher segmentation score according to existing CWS model, then it will have a higher probability in Eq.~(\ref{eq.Q1}), and we regularize our neural CWS model so that it tends to generate this tag sequence in Eq.~(\ref{eq.loss_PR}). 
In this way, the useful information in lexicon and unlabeled sentences can be incorporated into the learning of neural CWS model as indirect supervision.

Since a sentence usually has many possible tag sequences, following~\cite{zhang2017prior}, $\text{KL}$ function in Eq.~(\ref{eq.loss_PR}) is approximated as:
\begin{equation}
\label{eq.Q}
 \sum_{\mathbf{y}\in \mathcal{S}(\mathbf{x}_i)}\tilde{Q}(\mathbf{y}|\mathbf{x}_i;\boldsymbol{D},\boldsymbol{\theta}^t)\log(\frac{\tilde{Q}(\mathbf{y}|\mathbf{x}_i;\boldsymbol{D},\boldsymbol{\theta}^t)}{p(\mathbf{y}|\mathbf{x}_i;\boldsymbol{\theta})}),
\end{equation}
where $\mathcal{S}(\mathbf{x}_i)$ is a subset of $\mathcal{Y}(\mathbf{x}_i)$ whose elements have the highest prediction scores according to existing CWS model $\boldsymbol{\theta}^t$.
$\tilde{Q}$ is an approximation of $Q$ on the subset $\mathcal{S}(\mathbf{x}_i)$ as follows:
\begin{equation}
\tilde{Q}(\mathbf{y}|\mathbf{x}_i;\boldsymbol{D},\boldsymbol{\theta}^t)=\frac{\exp(\phi(\mathbf{y},\mathbf{x}_i;\boldsymbol{D},\boldsymbol{\theta}^t))}{\sum_{\hat{\mathbf{y}}\in\mathcal{S}(\mathbf{x}_i)}\exp( \phi(\hat{\mathbf{y}},\mathbf{x}_i;\boldsymbol{D},\boldsymbol{\theta}^t))}.
\end{equation}
In our approach, in each iteration the neural CWS model $\boldsymbol{\theta}$ is updated based on both labeled sentences and the indirect supervision from lexicon and unlabeled data.
The objective function for model update at iteration $t$ is formulated as follows:
\begin{equation}\label{eq.loss_all}
\small{
\mathcal{J}(\boldsymbol{\theta})=-\sum_{i=1}^{N_l}\log (p(\mathbf{y}_i|\mathbf{x}_i;\boldsymbol{\theta}))-
\lambda\sum_{i=1}^{N_u} \sum_{\hat{\mathbf{y}}\in\mathcal{S}(\mathbf{x}_i)} \tilde{Q}(\hat{\mathbf{y}}|\mathbf{x}_i;\boldsymbol{D},\boldsymbol{\theta}^t) \log(p(\hat{\mathbf{y}}|\mathbf{x}_i;\boldsymbol{\theta})),
}
\end{equation}
where $\lambda$ is a positive coefficient to control the relative importance of indirect supervision in model training.
In the first iteration, the initial neural CWS model $\boldsymbol{\theta}^1$ is trained on labeled sentences.
\section{Experiment}

\subsection{Datasets and Experimental Settings}

Two benchmark datasets released by the third international Chinese language processing bakeoff\footnote{{http://sighan.cs.uchicago.edu/bakeoff2006/download.html}}~\cite{levow2006third} are used in our experiments.
The first one is the \textit{MSRA} dataset, which contains 46,364 labeled sentences for training and 4,365 labeled sentences for test.
The second one is the \textit{UPUC} dataset, which contains 18,804 and 5,117 labeled sentences for training and test.


The lexicon used in our experiments is the Sogou Chinese lexicon\footnote{{http://www.sogou.com/labs/resource/list\_lan.php}}.
The size of character embeddings is 200.
These character embeddings are pretrained on the Sogou news corpus\footnote{{http://www.sogou.com/labs/resource/ca.php}} using the word2vec~\cite{mikolov2013efficient} tool.
We use 400 convolutional kernels in the CNN network, and the sizes of these kernels vary from 2 to 5.
$\lambda$ in Eq.~(\ref{eq.loss_all}) and $\alpha$ in Eq.~(\ref{eq.phi}) are set to 0.5 and 1 respectively.
We apply dropout technique to the embedding layer and the CNN layer, and the dropout rate is 0.3.
RMSProp~\cite{dauphin2015equilibrated} algorithm is used for model training.
The learning rate is 0.001 and the batch size is 64.
These hyper-parameters are selected using validation data.
Following~\cite{chen2015long}, we use the last 10\% sentences in the training set as validation data, and the remaining labeled sentences for model training.
In addition, we randomly sample 50\% training data as unlabeled data.
Each experiment is repeated 5 times and the average results are reported.

\subsection{Performance Evaluation}
In this section we evaluate the performance of our approach by comparing it with many baseline methods for Chinese word segmentation.
These methods include:
(1) LSTM-CRF, the most popular neural method for CWS based on the LSTM-CRF architecture~\cite{peng2017multi};
(2) CNN-CRF, the neural CWS method based on the CNN-CRF architecture, which is the basic model in our approach;
(3) Chen~\shortcite{chen2015long}, a neural CWS method using LSTM to learn character features and also considering local contexts;
(4) Zhang~\shortcite{zhang2018neural}, an LSTM-CRF based CWS method which integrates lexicon into model training via feature templates;
(5) Liu~\shortcite{liu2018neural}, a neural CWS method which incorporates lexicon into model training via multi-task learning;
(6) Liu~\shortcite{liu2014domain}, a CRF based CWS method which utilizes lexicon and unlabeled data to build partially labeled data for model training;
(7) Zhao~\shortcite{zhao2018neural}, an LSTM based CWS method which incorporates lexicon and unlabeled data via building partially labeled data;
(8) LUPR, our proposed neural CWS approach with both lexicon and unlabeled data via posterior regularization.
We conducted experiments on different ratios of training data, and the experimental results of different methods are summarized in Tables~\ref{result_msra} and~\ref{result_upuc}.

\begin{table}[!htb]
  \caption{The results on the \emph{MSRA} dataset. $P$, $R$ and $F$ represent precision, recall and Fscore respectively.}\label{result_msra}
  \centering
  \resizebox{0.45\textwidth}{!}{
    \begin{tabular}{@{\extracolsep{0pt}} c| c| c| c| c| c| c| c| c| c}
      \Xhline{1pt}
      \multirow{2}{*}{} &
      \multicolumn{3}{c|}{1\%} & \multicolumn{3}{c|}{5\%} & \multicolumn{3}{c}{10\%} \\
      \Xcline{2-10}{0.5pt}
      & $P$ & $R$  & $F$ & $P$ & $R$  & $F$ & $P$ & $R$  & $F$ \\
      \Xhline{0.9pt}
      LSTM-CRF  & 75.87 & 76.18 & 76.01 & 82.81 & 82.18 & 82.49 & 85.24 & 84.68 & 84.95 \\
      \hline
      CNN-CRF  & 77.19 & 75.35 & 76.26 & 87.19 & 86.64 & 86.91 & 89.95 & 89.48 & 89.71 \\
      \hline
      Chen~\cite{chen2015long}  & 77.20 & 74.32 & 75.73 & 84.19 & 83.44 & 83.80 & 87.50 & 86.05 & 86.76 \\
      \hline
      \hline
      Zhang~\cite{zhang2018neural}  & 76.64 & 76.55 & 76.60 & 87.15 & 86.73 & 86.94 & 89.49 & 89.10 & 89.29 \\
      \hline
      Liu~\cite{liu2018neural} & 78.06 & 77.55 & 77.80 & 87.60 & 86.50 & 87.05 & 90.06 & 89.47 & 89.77 \\
      \hline
      \hline
      Liu~\cite{liu2014domain}  & 81.48 & 78.92 & 80.18 & 83.76 & 81.58 & 82.66 & 85.20 & 83.09 & 84.13 \\
      \hline
      Zhao~\cite{zhao2018neural}  & 80.68 & \textbf{80.26} & 80.47 & 86.94 & 85.67 & 86.30 & 88.69 & 87.21 & 87.94 \\
      \hline
      \hline
      LUPR  & \textbf{81.98} & 79.74 & \textbf{80.84} & \textbf{88.42} & \textbf{87.92} & \textbf{88.17} & \textbf{90.35} & \textbf{89.83} & \textbf{90.09} \\
      \Xhline{1pt}
    \end{tabular}
    }
\end{table}

\begin{table}[!htb]
  \caption{The results on the \emph{UPUC} dataset.}\label{result_upuc}
  \centering
  \resizebox{0.45\textwidth}{!}{
  \begin{tabular}{@{\extracolsep{0pt}} c| c| c| c| c| c| c| c| c| c}
      \Xhline{1pt}
      \multirow{2}{*}{} &
      \multicolumn{3}{c|}{1\%} & \multicolumn{3}{c|}{5\%} & \multicolumn{3}{c}{10\%} \\
      \Xcline{2-10}{0.5pt}
      & $P$ & $R$  & $F$ & $P$ & $R$  & $F$ & $P$ & $R$  & $F$ \\
      \Xhline{0.9pt}
      LSTM-CRF  & 70.49 & 73.44 & 71.92 & 79.66 & 80.95 & 80.29 & 82.97 & 84.73 & 83.84 \\
      \hline
      CNN-CRF  & 72.03 & 74.50 & 73.22 & 82.60 & 84.23 & 83.40 & 87.75 & 88.79 & 88.27 \\
      \hline
      Chen~\cite{chen2015long}  & 73.04 & 74.24 & 73.61 & 80.87 & 81.56 & 81.20 & 85.03 & 87.08 & 86.04 \\
      \hline
      \hline
      Zhang~\cite{zhang2018neural}  & 74.93 & 73.45 & 74.18 & 84.39 & 85.94 & 85.15 & 88.15 & 89.05 & 88.60 \\
      \hline
      Liu~\cite{liu2018neural} & 73.01 & 76.33 & 74.63 & 83.82 & 85.70 & 84.75 & 87.71 & 89.27 & 88.48 \\
      \hline
      \hline
      Liu~\cite{liu2014domain}  & \textbf{79.50} & 75.65 & 77.53 & 81.83 & 78.25 & 80.00 & 83.39 & 80.22 & 81.77 \\
      \hline
      Zhao~\cite{zhao2018neural}  & 77.11 & 77.57 & 77.34 & 82.70 & 82.07 & 82.39 & 85.19 & 86.07 & 85.63 \\
      \hline
      \hline
      LUPR  & 78.18 & \textbf{78.49} & \textbf{78.33} & \textbf{86.89} & \textbf{87.36} & \textbf{87.13} & \textbf{89.48} & \textbf{90.29} & \textbf{89.88} \\
      \Xhline{1pt}
    \end{tabular}
    }
\end{table}

According to Tables~\ref{result_msra} and~\ref{result_upuc}, our approach can outperform many neural Chinese word segmentation methods such as CNN-CRF, LSTM-CRF and Chen~\cite{chen2015long}.
In addition, the advantage of our approach over these baseline methods becomes bigger when the number of training samples decreases.
This is because these methods rely on a large number of labeled sentences to train neural CWS models, and cannot exploit the useful information in lexicon and unlabeled data.
When training data is insufficient, it is very difficult for these methods to train accurate and robust CWS models.
Since our approach can exploit the useful information in both lexicon and unlabeled data, it can reduce the dependence on labeled sentences and achieve better performance than these baseline methods.

Although the methods proposed in \cite{zhang2018neural} and \cite{liu2018neural} can also incorporate lexicon information into neural CWS model training, our approach can consistently outperform them.
In \cite{zhang2018neural} the lexicon is utilized via handcrafted feature templates, which need a lot of domain knowledge to design and may not be optimal.
In addition, an extra LSTM network is incorporated to learn hidden character representations from these lexicon based features, making the neural model more difficult to train when labeled data is scarce.
In \cite{liu2018neural} the lexicon is utilized via an auxiliary word classification task which is jointly trained with CWS model.
Although these methods can utilize the lexicon for neural CWS, the useful information in massive unlabeled data is not considered.
Our approach can exploit both lexicon and unlabeled data for neural Chinese word segmentation.
Thus, our approach can consistently outperform them.

Although the methods in \cite{liu2014domain} and \cite{zhao2018neural} can also exploit both lexicon and unlabeled data for CWS, our approach can still outperform them.
In \cite{liu2014domain} and \cite{zhao2018neural}, the lexicon and unlabeled data are used to build partially annotated datasets.
However, the partially annotated data constructed by word matching based on lexicons may contain heavy noise and the context information of sentences is not considered.
In our approach, the lexicon and unlabeled data are used to provide indirect supervision for model training by regularizing the prediction space of the neural CWS model.
The experimental results show that our approach is more effective in exploiting Chinese lexicon and unlabeled data for CWS than \cite{liu2014domain} and \cite{zhao2018neural}.

\subsection{Effect of Lexicon and Unlabeled Data}

\begin{figure}[!t]
  \centering
  \subfigure[\textit{MSRA} dataset.]{
    \label{fig.msra_dtype} 
    \includegraphics[width=0.45\linewidth]{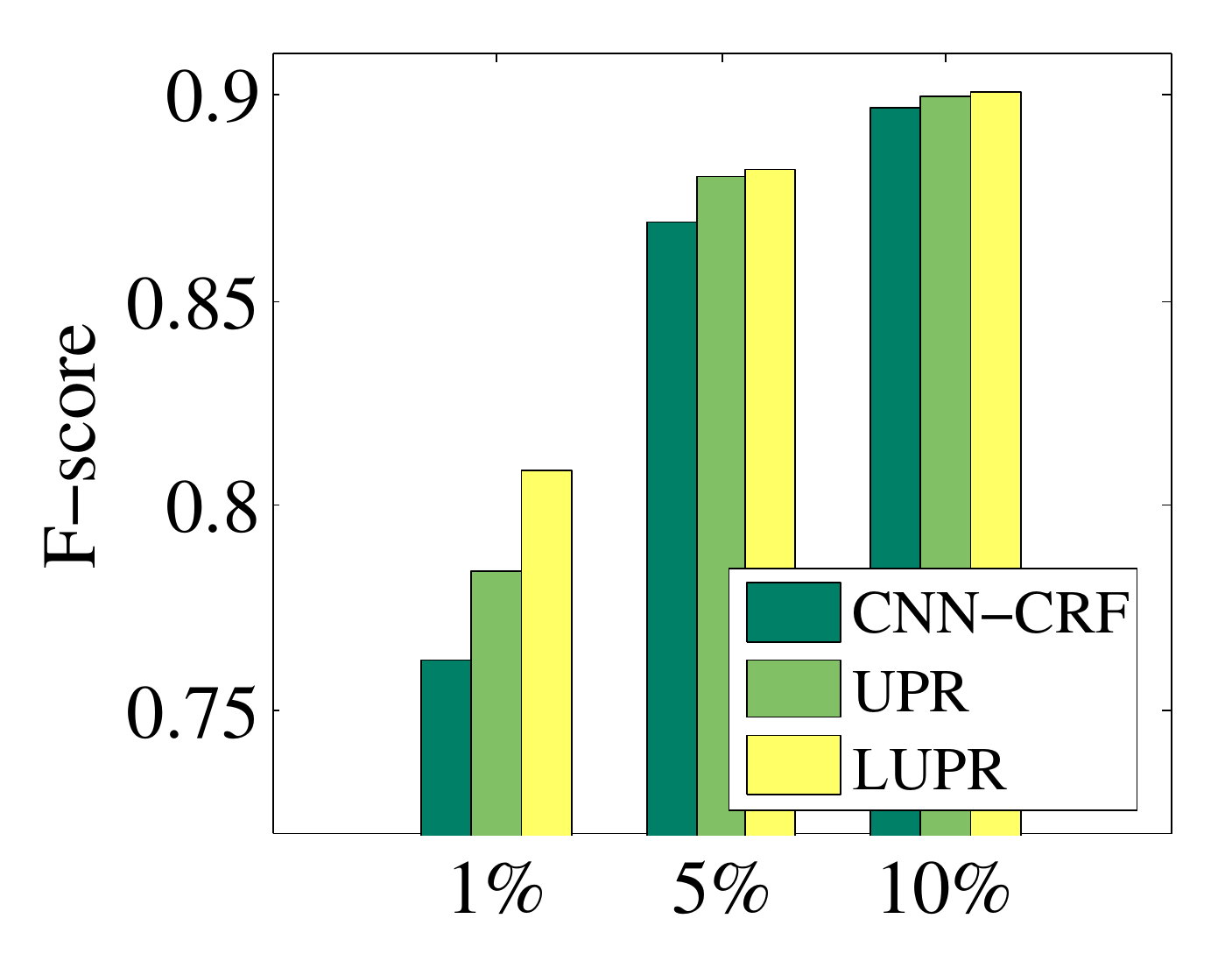}}
  \hspace{0in}
  \subfigure[\textit{UPUC} dataset.]{
    \label{fig.upuc_dtype} 
    \includegraphics[width=0.45\linewidth]{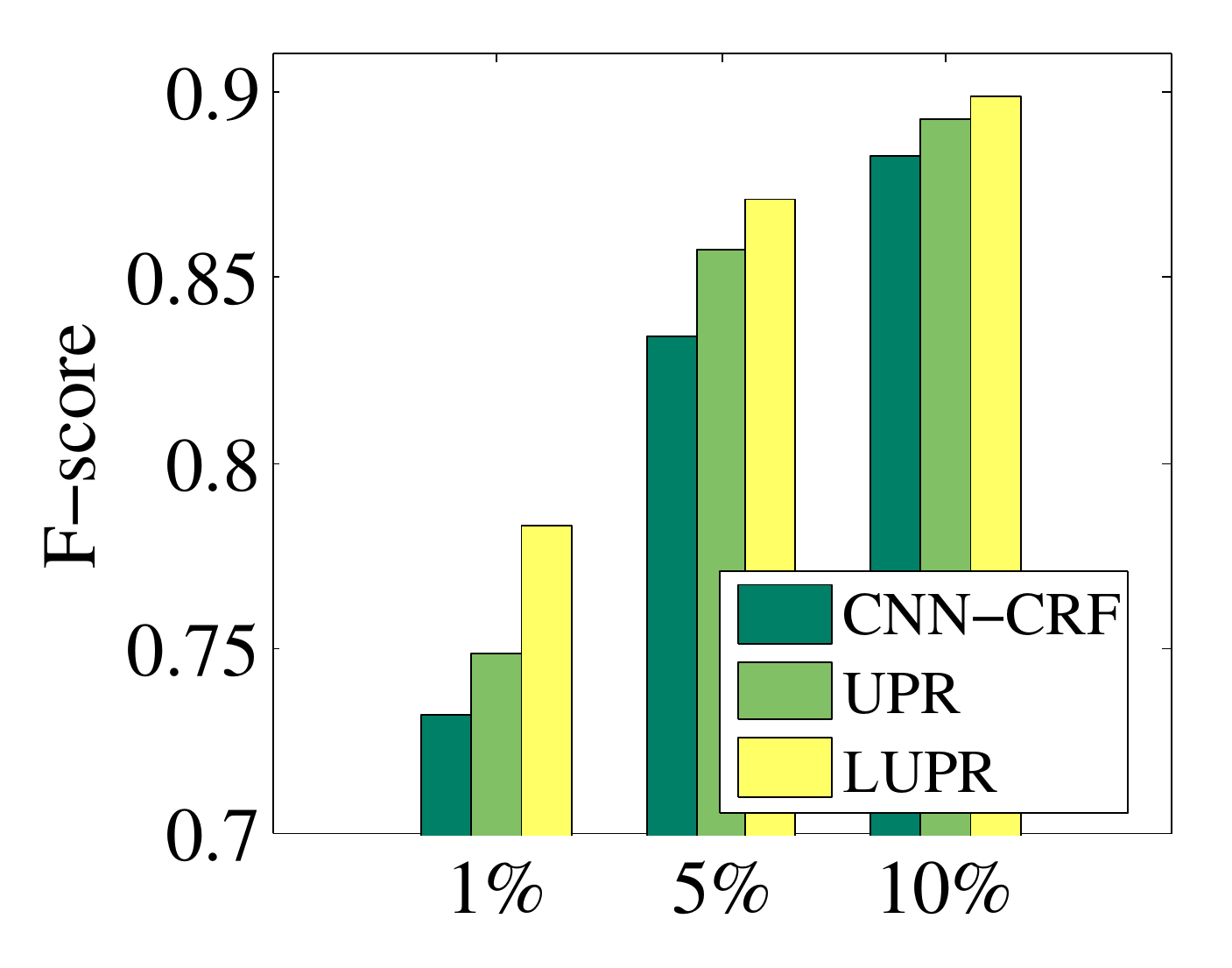}}
  \vspace{-0.05in}
  \caption{The performance of the basic CNN-CRF model and our approach with only unlabeled data (UPR) and with both lexicon and unlabeled data (LUPR).}
  \label{fig.dtype}
  \vspace{-0.05in}
\end{figure}
In this section, we conducted experiments to explore the effectiveness of lexicon and unlabeled data for neural Chinese word segmentation.
The experimental results are summarized in Fig.~\ref{fig.dtype}.

According to Fig.~\ref{fig.dtype}, incorporating unlabeled data can effectively improve the performance of CWS in our approach.
This is because the massive unlabeled data contains rich useful information for word segmentation.
For example, if the character sequence ``特朗普'' frequently appear in unlabeled sentences with different contexts, then we can infer that it is probably a unique Chinese word.
The result in Fig.~\ref{fig.dtype} shows that our approach is effective in exploiting the useful information in unlabeled data for CWS.
In addition, according to Fig.~\ref{fig.dtype} after incorporating lexicon the performance of our approach can be further improved.
This is because the Chinese lexicon can provide important information of whether a character sequence can be a valid Chinese word, which is useful for the CWS task.
The result in Fig.~\ref{fig.dtype} validates that our approach can effectively exploit Chinese lexicon to improve the performance of neural CWS.

\subsection{Size of Lexicon and Unlabeled Data}
\begin{figure}[!t]
  \centering
  \subfigure[Ratio of unlabeled data.]{
    \label{fig.ulsize} 
    \includegraphics[width=0.45\linewidth]{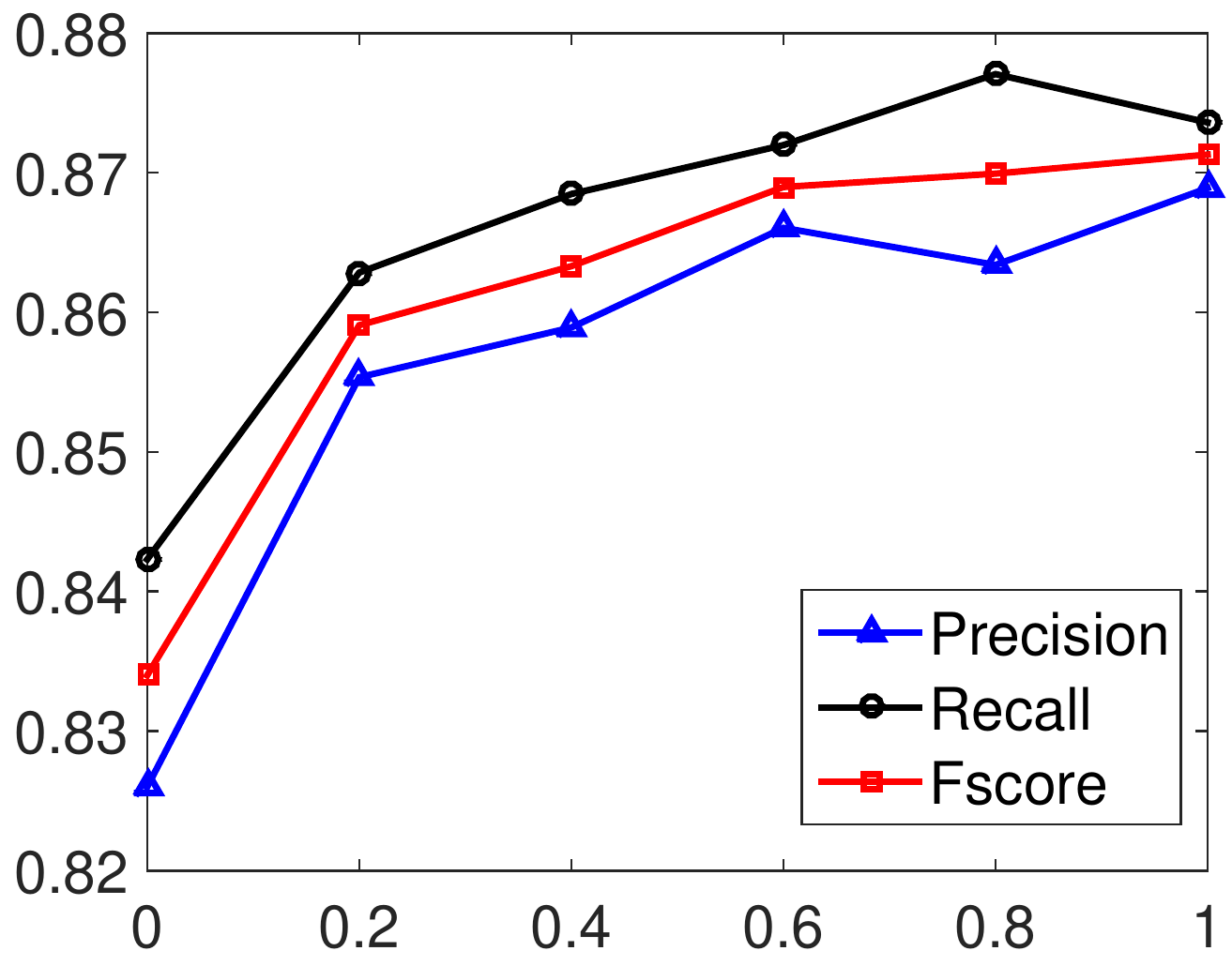}}
  \hspace{0in}
  \subfigure[Ratio of words in lexicon.]{
    \label{fig.dsize} 
    \includegraphics[width=0.45\linewidth]{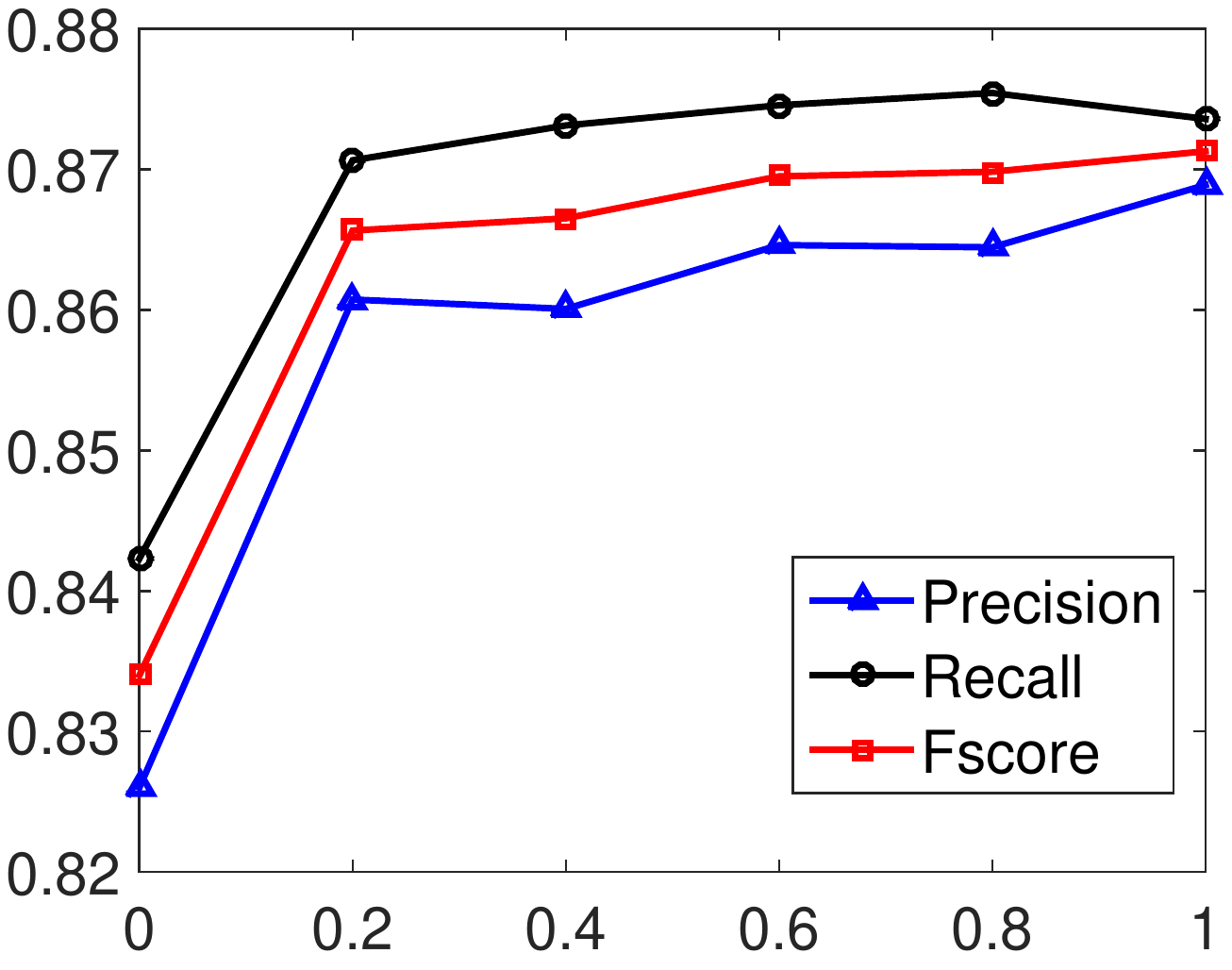}}
  \vspace{-0.05in}
  \caption{Influence of unlabeled data and lexicon size.}
  \label{fig.size}
  \vspace{-0.05in}
\end{figure}

In this section, we conduct experiments to explore the influence of the sizes of unlabeled data and Chinese lexicon on the performance of our approach.
The experiments were conducted on the \emph{UPUC} dataset, and we randomly sampled 5\% of the training data for model training.
The experiments on the \emph{MSRA} dataset show similar patterns.
The experimental results of unlabeled data size are summarized in Fig.~\ref{fig.ulsize}.
We can see that as more unlabeled data is incorporated, the performance of our approach consistently improves.
This result further validates that the unlabeled data contains a lot of useful information for Chinese word segmentation, and our approach is effective in exploiting unlabeled data for neural CWS methods.
The experimental results of lexicon size are shown in Fig.~\ref{fig.dsize}.
According to Fig.~\ref{fig.dsize}, as more Chinese words are included in the lexicon, the performance of our approach improves.
This is because with more Chinese words in the lexicon, our approach can have better capacity in recognizing the boundaries of words which rarely or never appear in training data but are included in the Chinese lexicon.

\subsection{Influence of Hyper-parameters}

\begin{figure}[!t]
  \centering
  \subfigure[\textit{MSRA} dataset.]{
    \label{fig:self} 
    \includegraphics[width=0.45\linewidth]{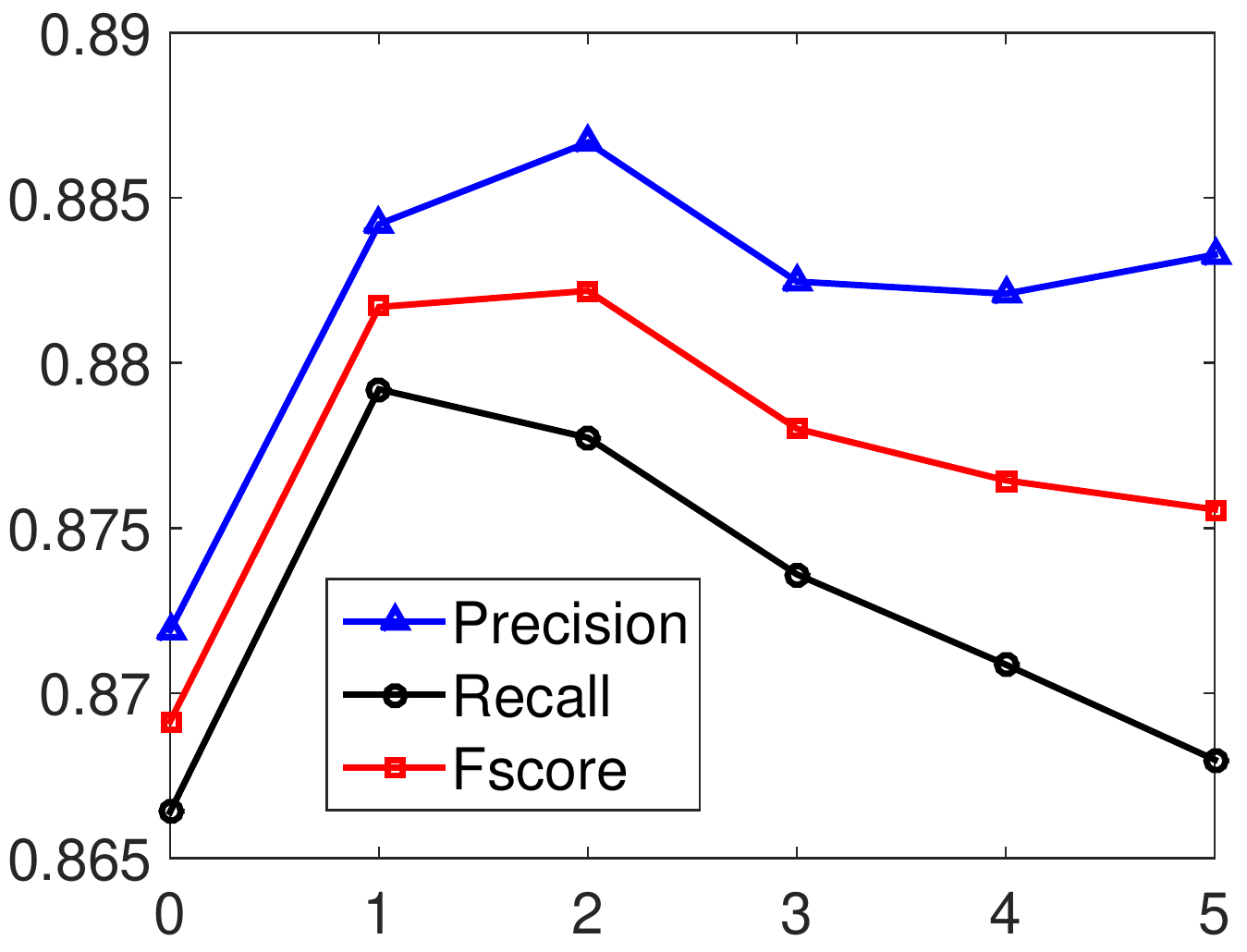}}
  \hspace{0in}
  \subfigure[\textit{UPUC} dataset.]{
    \label{fig:self2} 
    \includegraphics[width=0.45\linewidth]{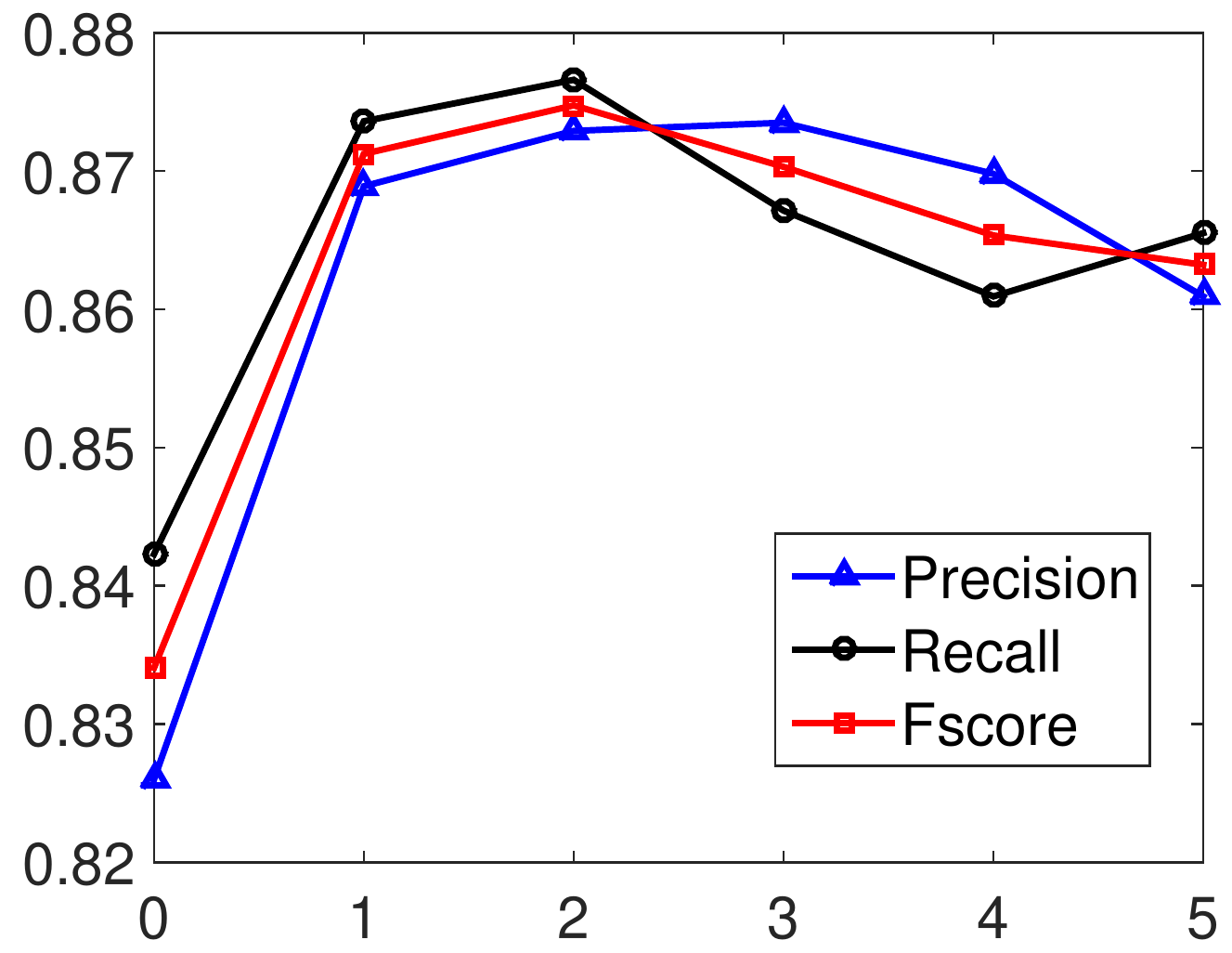}}
  \vspace{-0.05in}
  \caption{Influence of $S$ size in Eq.~(\ref{eq.Q}).}
  \label{fig.SN}
  \vspace{-0.05in}
\end{figure}

\begin{figure}[!t]
  \centering
  \subfigure[\textit{MSRA} dataset.]{
    \label{fig:self} 
    \includegraphics[width=0.45\linewidth]{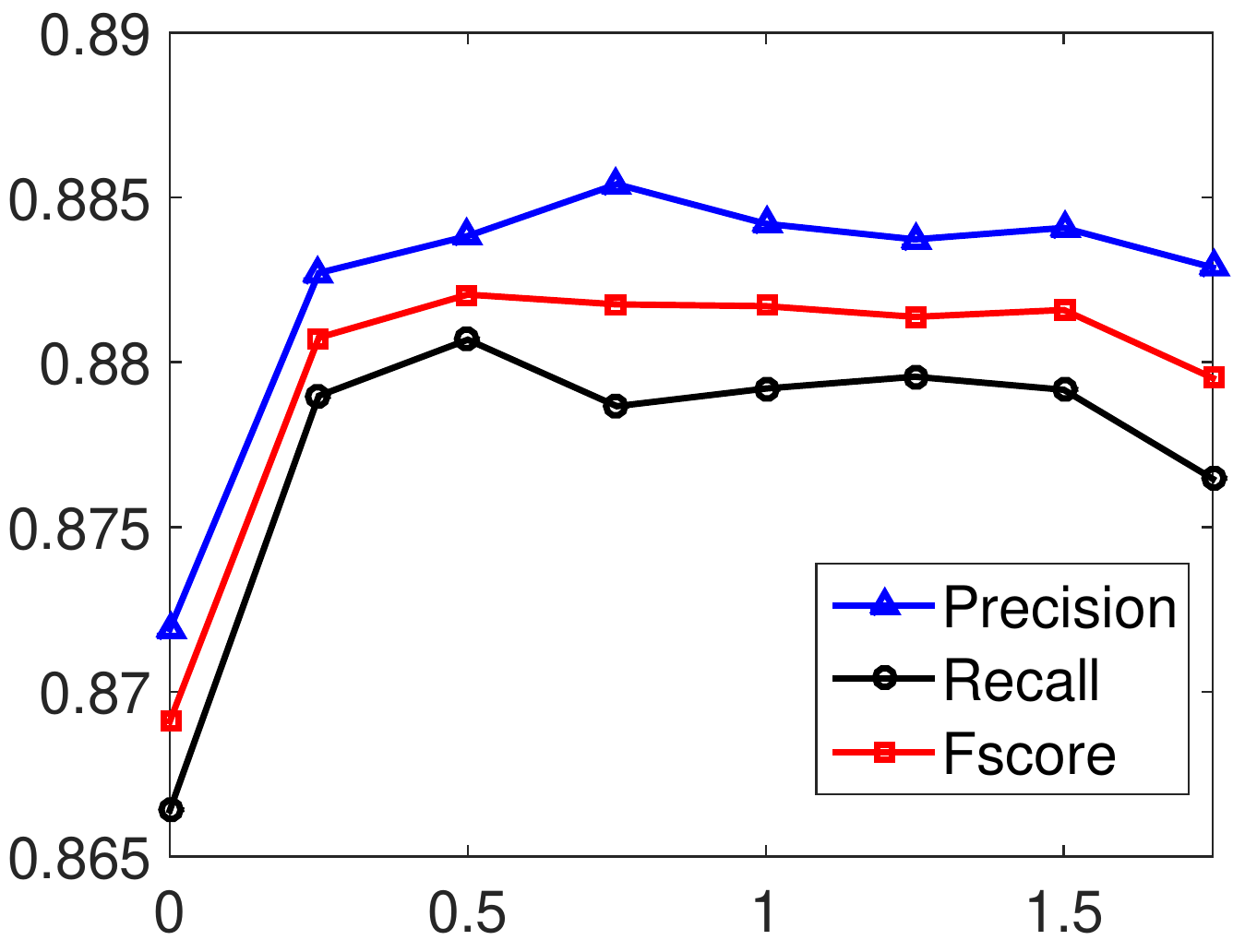}}
  \hspace{0in}
  \subfigure[\textit{UPUC} dataset.]{
    \label{fig:self2} 
    \includegraphics[width=0.45\linewidth]{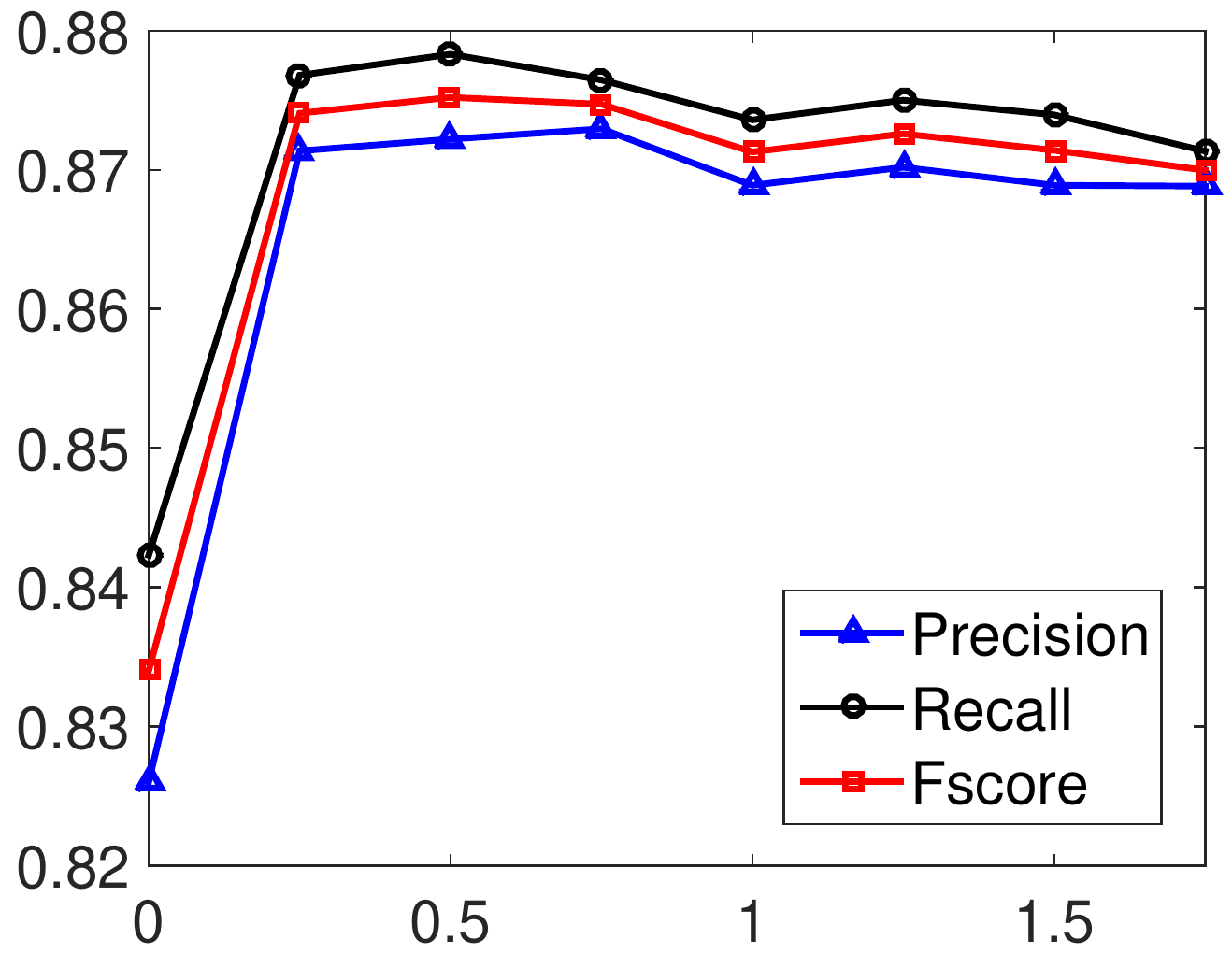}}
  \vspace{-0.05in}
  \caption{Influence of $\lambda$ in Eq.~(\ref{eq.loss_all}).}
  \label{fig.lambda}
  \vspace{-0.05in}
\end{figure}

\begin{figure}[!t]
  \centering
  \subfigure[\textit{MSRA} dataset.]{
    \label{fig:self} 
    \includegraphics[width=0.45\linewidth]{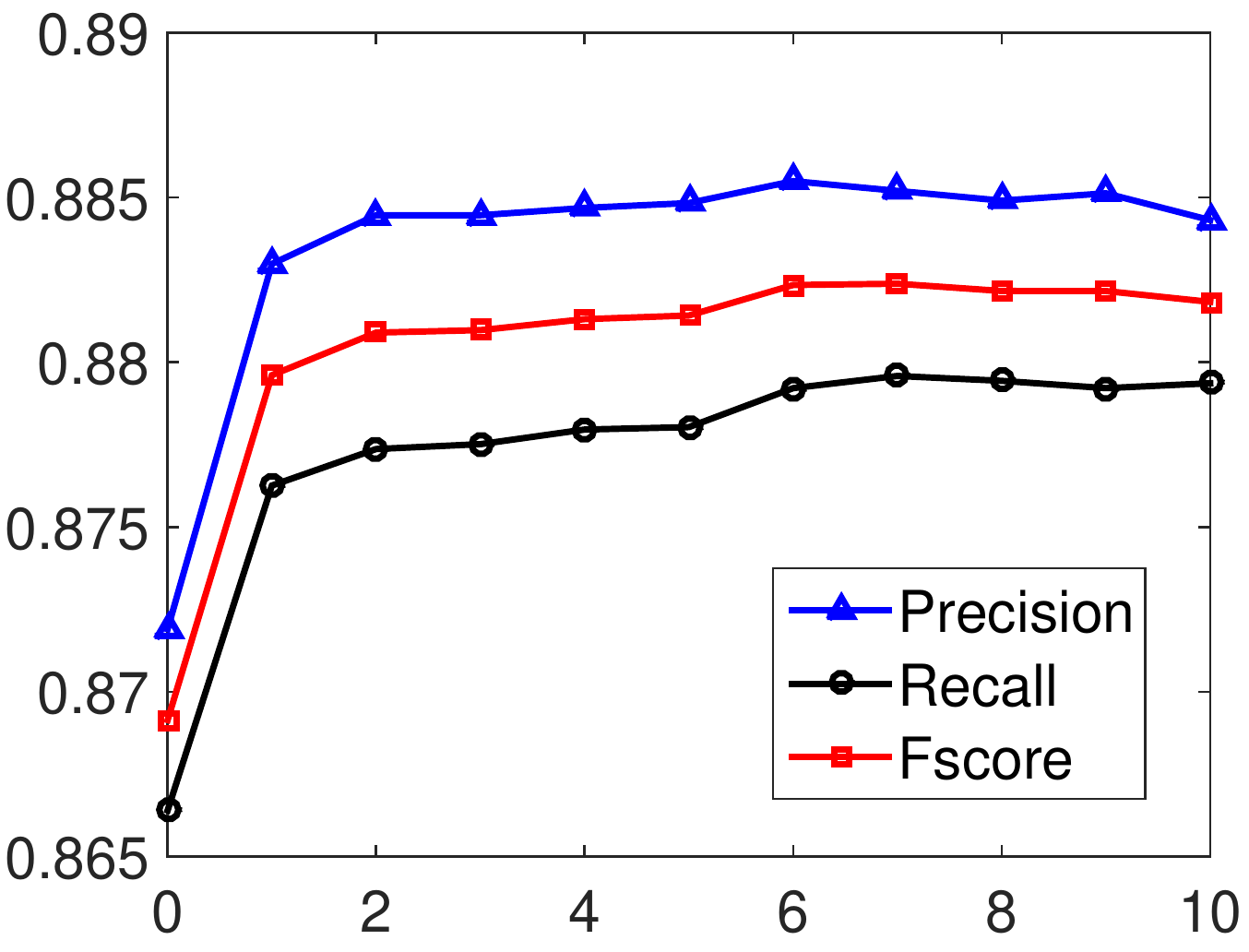}}
  \hspace{0in}
  \subfigure[\textit{UPUC} dataset.]{
    \label{fig:self2} 
    \includegraphics[width=0.45\linewidth]{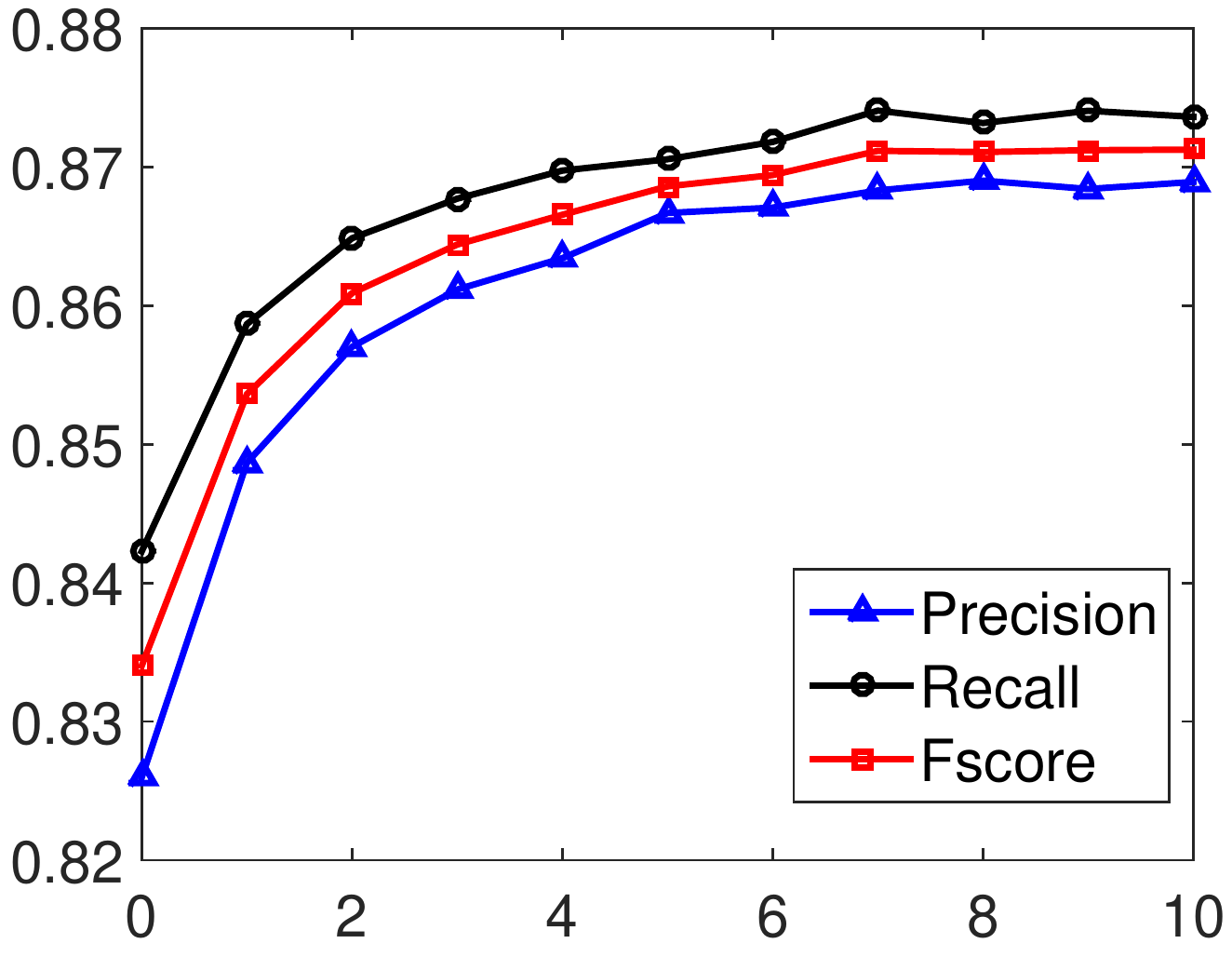}}
  \vspace{-0.05in}
  \caption{Influence of iteration number.}
  \label{fig.iternum}
  \vspace{-0.05in}
\end{figure}

In this section, we conducted experiments to explore the influence of three important hyper-parameters, i.e., the size of set $S$ in Eq.~(\ref{eq.Q}), $\lambda$ in Eq.~(\ref{eq.loss_all}) and iteration number, on the performance of our approach.
The experimental results of the size of set $S$ in Eq.~(\ref{eq.Q}) are shown in Fig.~\ref{fig.SN}.
In these experiments, we randomly sampled 5\% of the training data for model training.
According to Fig.~\ref{fig.SN}, when the size of $S$ in Eq.~(\ref{eq.Q}) increases from 0 to 1, the performance of our approach significantly improves.
This is because when the size of $S$ is 0, the useful information in unlabeled data and lexicon is not incorporated.
Thus, the performance is not optimal.
When the size of $S$ becomes too large, the performance of our approach slightly decreases.
It indicates that the number of informative tag sequences of unlabeled sentences inferred from lexicon is usually limited, and incorporating too many of them may introduce some noisy information.
A moderate size of $S$ (e.g., 2) is most suitable for our approach.

The experimental results of $\lambda$ are shown in Fig.~\ref{fig.lambda}.
$\lambda$ is used to control the relative importance of the indirect supervision from unlabeled data and lexicon in Eq.~(\ref{eq.loss_all}).
According to Fig.~\ref{fig.lambda}, as $\lambda$ increases, the performance of our approach first increases and then slightly decreases.
This is because when $\lambda$ is too small, the useful information in the lexicon and unlabeled data is not fully exploited.
Thus, the performance is not optimal.
However, when $\lambda$ becomes too large, the indirect supervision inferred from lexicon and unlabeled data is over-emphasized and the labeled sentences are not fully respected.
Thus, the performance starts to decline.
A moderate value of $\lambda$ is most appropriate for our approach.

The experimental results of iteration number are shown in Fig.~\ref{fig.iternum}. 
According to Fig.~\ref{fig.iternum}, as the iteration number grows, the performance of our approach first improves and then gradually becomes stable.
This is because in each iteration the neural CWS model can be enhanced by incorporating the unlabeled data and lexicon, and the refined CWS model in turn can help improve the indirect supervision in the next iteration of our approach.
This result validates the effectiveness of our approach in iteratively training neural CWS model by exploiting both labeled sentences and the indirect supervision inferred from the unlabeled sentences and Chinese lexicons.

\subsection{Domain Adaptation for CWS}
In Chinese word segmentation field, several domains (e.g., news) have accumulated much labeled data, while in many other domains (e.g., medical records) labeled data for CWS is scarce and even nonexistent.
Although annotating sufficient labeled data for these domains is time-consuming and expensive, the unlabeled sentences are usually easy to collect.
In addition, in many target domains there are off-the-shelf lexicons or it is relatively easy to build one.
Thus, an interesting application of our approach is domain adaptation for Chinese word segmentation, i.e., using labeled sentences in a source domain (e.g., news) and unlabeled sentences in a target domain (e.g., medical records) as well as the lexicon from the target domain to train a robust neural CWS model for target domain.
In this section we conduct experiments to explore the performance of our approach in domain adaptation for word segmentation.

Two datasets are used in our experiments.
The first one is the \emph{Zhuxian} dataset\footnote{{http://zhangmeishan.
github.io/eacl14mszhang.zip}} built by Zhang et al.~\shortcite{zhang2014type} from a Chinese online novel.
We used the same lexicon\footnote{This lexicon is crawled from \url{http://baike.baidu.com/view/18277.htm}} as~\cite{zhao2018neural} for this domain.
In addition, we used the same unlabeled data as~\cite{liu2014domain} which contains about 16K sentences.
The second dataset is the Weibo dataset\footnote{{https://github.com/FudanNLP/NLPCC-WordSeg-Weibo}} released by the Weibo word segmentation task of NLPCC2016~\cite{qiu2016overview}.
This dataset contains 20,135 sentences for training and 8,592 sentences for test.
Since there is no off-the-shelf lexicon for Weibo word segmentation, we built one using the words extracted from the training data.
In addition, we regarded the training sentences as unlabeled data.
We used \emph{Zhuxian} and \emph{Weibo} datasets as two target domains, and used the \emph{UPUC} dataset as source domain.
The experimental results are summarized in Table~\ref{table.result_domain}.
The settings of our approach and baseline methods are consistent with previous experiments.

\begin{table}[htbp]
\caption{Experimental results of domain adaptation.}
\vspace{-0.05in}
  \centering
  \resizebox{0.4\textwidth}{!}{
  \begin{tabular}{@{\extracolsep{0pt}} c|c|c|c|c|c|c}
  \Xhline{1pt}
  & \multicolumn{3}{c|}{Zhuxian} & \multicolumn{3}{c}{Weibo}\\
  \cline{2-7}
  & $P$ & $R$ & $F$ & $P$ & $R$ & $F$ \\
  \hline
  CNN-CRF & 87.95 & 87.68 & 87.81 & 86.11 & 88.23 & 87.16\\
  \hline
  LSTM-CRF & 85.94 & 86.26 & 86.10 & 86.39 & 88.35 & 87.36\\
  \hline
  Chen~\cite{chen2015long} & 85.24 & 87.57 & 86.39 & 86.87 & 89.28 & 88.06\\
  \hline
  \hline
  Zhang~\cite{zhang2018neural} & 88.70 & 87.89 & 88.29 & 88.94 & \textbf{90.89} & 89.90\\
  \hline
  \hline
  Liu~\cite{liu2018neural} & 87.79 & 88.04 & 87.91 & 86.24 & 88.49 & 87.35\\
  \hline
  \hline
  Liu~\cite{liu2014domain} & 89.86 & 89.04 & 89.44 & \textbf{90.31} & 89.89 & 90.10\\
  \hline
  Zhao~\cite{zhao2018neural} & 88.50 & \textbf{91.32} & 89.90 & 89.83 & 85.07 & 87.39\\
  \hline
  \hline
  LUPR & \textbf{90.53} & 89.39 & \textbf{89.95} & 89.85 & 90.73 & \textbf{90.29}\\
  \Xhline{1pt}
  \end{tabular}
}
\label{table.result_domain}
\end{table}

According to Table~\ref{table.result_domain}, the performance of neural models that are trained on labeled data of source domain such as CNN-CRF, LSTM-CRF and \cite{chen2015long} is relatively low in target domains.
This is because there is huge difference in word distribution between source and target domains.
Many words in target domain may not or rarely appear in the labeled data of source domain, making it difficult for these CWS methods to segment the sentences in target domain.
The methods proposed in \cite{zhang2018neural} and \cite{liu2018neural} which incorporate lexicon can effectively improve the performance of CWS in target domains.
Our approach can outperform these methods because beyond the lexicon information our approach can also incorporate massive unlabeled data into neural model training, which can provide useful information for CWS.
Although \cite{liu2014domain} and \cite{zhao2018neural} can incorporate both lexicon and unlabeled data for word segmentation, our approach can outperform them in the domain adaptation scenario.
This result implies that incorporating the lexicon and unlabeled data as indirect supervision via posterior regularization is more suitable for domain adaptation of CWS than utilizing them to build partially labeled data through word matching, which may introduce heavy noise.

\subsection{Case Study}

In this section, we conducted several case studies to further explore why our approach can improve the performance of CWS.
We studied several cases in the domain adaptation experiments, where the source domain is \emph{UPUC} and the target domain is \emph{Weibo}. 
Several examples are illustrated in Table~\ref{case}.

\begin{table}[h]
\caption{Several examples of Chinese word segmentation.}
\vspace{-0.05in}
  \centering
  \resizebox{0.4\textwidth}{!}{
  \begin{tabular}{@{\extracolsep{0pt}} c|c|c}
    \Xhline{1pt}
     & Example 1 & Example 2 \\
    \hline
    Sentence & 养老金调整三大焦点问题 & 穆勒操刀梅开二度 \\
    \hline 
    \hline 
    CNN-CRF & 养/老金/调整/三/大/焦点/问题 & 穆勒/操刀/梅开/二/度 \\
    \hline
    LUPR & 养老金/调整/三/大/焦点/问题 & 穆勒/操刀/梅开二度 \\
  \Xhline{1pt}
\end{tabular}
}
\label{case}
\vspace{-0.05in}
\end{table}

According to Table~\ref{case}, our approach performs better than baseline methods in domain adaptation scenario, especially on the sentences with OOV and rare words of source domain.
For example, in the first example ``养老金" is a rare word in the training data of source domain, and in the second example ``梅开二度" is an OOV word in source domain.
The segmentation results of the basic CNN-CRF model on these words are not correct.
Since ``养老金" is a popular word in the unlabeled data of target domain and ``梅开二度" is included in the target domain lexicon, our approach can correctly segment these sentences.
Thus, our approach can improve the performance of Chinese word segmentation by exploiting the useful information in both lexicon and unlabeled data.

\section{Conclusion}

In this paper, we propose a neural approach for Chinese word segmentation which can exploit the useful information in both Chinese lexicon and unlabeled sentences for model training.
Our approach is based on the posterior regularization algorithm, and we propose a unified framework to incorporate both unlabeled data and lexicon to provide indirect supervision for model training by regularizing the prediction space of the neural CWS models.
Extensive experiments are conducted on multiple benchmark datasets in both in-domain and cross-domain scenarios.
The experimental results validate that our approach can effectively improve the performance of neural Chinese word segmentation.
\begin{acks}
This work was supported in part by the National Key Research and Development Program of China under
Grant 2018YFC1604002 and in part by the National Natural Science Foundation of China under Grant U1705261, Grant U1536201, Grant U1536207.
\end{acks}
\end{CJK*}

\bibliographystyle{ACM-Reference-Format}
\balance 
\bibliography{main}

\end{document}